\documentclass[sigconf]{acmart}
\usepackage{algorithm}
\usepackage{algorithmic}
\usepackage{multirow}
\AtBeginDocument{%
  }

\copyrightyear{2026}
\acmYear{2026}
\setcopyright{cc}
\setcctype{by}
\acmConference[BCB '26]{17th ACM International Conference on Bioinformatics, Computational Biology and Health Informatics}{June 30-July 03, 2026}{Rende (CS), Italy}
\acmBooktitle{17th ACM International Conference on Bioinformatics, Computational Biology and Health Informatics (BCB '26), June 30-July 03, 2026, Rende (CS), Italy}
\acmDOI{10.1145/3807503.3819375}
\acmISBN{979-8-4007-2653-8/2026/06}

\begin{document}

\title{LTR-ICD: A Ranking-Aware Framework for Automatic ICD Coding}

\author{Mohammad Mansoori}
\affiliation{%
  \institution{Halmstad University}
  \city{Halmstad}
  \state{Halland}
  \country{Sweden}
}
\email{mohammad.mansoori@hh.se}
\orcid{0009-0003-9554-4675}

\author{Amira Soliman}
\affiliation{%
  \institution{Halmstad University}
  \city{Halmstad}
  \state{Halland}
  \country{Sweden}
}
\email{amira.soliman@hh.se}
\orcid{0000-0002-0264-8762}

\author{Farzaneh Etminani}
\affiliation{%
  \institution{Halmstad University}
  \city{Halmstad}
  \state{Halland}
  \country{Sweden}
}
\email{farzaneh.etminani@hh.se}
\orcid{0000-0003-2006-6229}

\renewcommand{\shortauthors}{Mansoori et al.}

\begin{abstract}
  Clinical notes contain unstructured text provided by clinicians during patient encounters. These notes are usually accompanied by a sequence of diagnostic codes following the International Classification of Diseases (ICD). Correctly assigning and ordering ICD codes is essential for medical diagnosis and reimbursement. However, automating this task remains challenging. State-of-the-art methods treated this problem as a classification task, leading to ignoring the order of ICD codes that is essential for different purposes. In this work, as a first attempt, we approach this task from a retrieval system perspective to consider the order of codes, thus formulating this problem as a classification and ranking task. Our results and analysis show that the proposed framework has a superior ability to identify high-priority codes compared to other methods. For instance, our model’s accuracy in correctly ranking primary diagnosis codes is \textasciitilde47\%, compared to \textasciitilde20\% for the state-of-the-art classifier. Additionally, in terms of classification metrics, the proposed model achieves a micro- and macro-F1 scores of 0.6065 and 0.2904, respectively, surpassing the previous best model with scores of 0.6035 and 0.2741.
\end{abstract}

\ccsdesc[500]{Applied computing~Health informatics}
\ccsdesc[500]{Computing methodologies~Natural language processing}
\ccsdesc[500]{Computing methodologies~Supervised learning by classification}
\ccsdesc[500]{Information systems~Language models}
\ccsdesc[500]{Information systems~Recommender systems}


\keywords{Generative Language Models, Rank Aware Classification, Automatic Medical Coding, ICD Coding, Electronic Health Records, Pre-trained Language Models}


\maketitle

\section{Introduction}
International Classification of Diseases (ICD) codes are alphanumeric codes used to classify diagnoses, symptoms, procedures, and other health conditions. These codes are part of the ICD system, maintained by the World Health Organization (WHO). This coding system follows a hierarchical structure, which organizes diseases and health conditions into chapters, categories, sub-categories, and codes. Medical coders or clinical documentation specialists usually assign ICD codes. These professionals examine patients' electronic health records (EHRs), including clinical notes, lab results, and other relevant documentation, to assign the appropriate ICD codes for the documented diagnoses and procedures. They ensure the codes are arranged in the proper order, a process known as ‘sequencing’, which is essential for accurate coding\footnote{\url{https://stacks.cdc.gov/view/cdc/126426}}.

The order of ICD codes can be crucial in many situations, particularly in certain contexts such as medical billing \cite{burns2012systematic}. For instance, when submitting claims to insurance companies for reimbursement, the primary diagnosis code, which represents the main reason for the patient's encounter with the healthcare provider, is typically listed first. Secondary diagnosis codes may follow, indicating additional conditions relevant to the patient's treatment or care \cite{OMalley2005}. Similarly, researchers and public health officials use ICD codes to analyze disease patterns, track trends, and evaluate the effectiveness of healthcare interventions \cite{gianfrancesco2021narrative}. The order of codes can affect the accuracy of these analyses and the validity of research findings.

Most recent studies formulate the task of automatic ICD coding as an extreme multi-label classification assignment, which learns the representations of EHR clinical notes with a deep learning-based encoder and predicts codes with a multi-label classifier \cite{9705116,10.1145/3664615}. However, these classifier models are typically designed to make binary decisions (e.g., whether an item belongs to a particular class or not), which can be limiting for predictive ICD coding. While ICD coding is fundamentally a classification task, in practical settings it also involves sequencing codes based on clinical priority and relevance. Standard classification models do not explicitly account for this ordering, which may reduce their effectiveness in real-world applications. Furthermore, the inherent complexity and variability of clinical notes make it difficult for the current models to achieve the accuracy and reliability required for a real-world automated ICD coding system.

Addressing the above-mentioned limitations, this research re-frames the problem of predictive ICD coding as a recommendation task. By formulating it this way, we aim to develop tools that can effectively assist human coders, integrating the strengths of both machine intelligence and human expertise. In essence, this study addresses the challenges of existing classifier models and their evaluation metrics for real-world applications. It aligns with the practical needs of healthcare providers, ensuring that ICD coding can be more accurately and efficiently conducted in real-world settings.

\textbf{Key Contributions:} As the first contribution, and to the best of our knowledge, this study is the first to formulate the problem of automatic ICD coding as a combined classification and ranking task. This joint formulation addresses key shortcomings of existing models and their evaluation metrics, enabling more accurate and practical ICD code assignment in real-world clinical settings. As the second contribution, we propose LTR-ICD, a novel language model-based framework that recommends order-aware ICD codes for input clinical notes. This new architecture includes a classification module and a generative module with a shared encoder, trained jointly through a two-stage learning process. Experimental results on the benchmark MIMIC-III dataset demonstrate that the proposed framework significantly improves both the ranking quality of the recommended labels and their classification performance compared to previous state-of-the-art classifier models.

\section{Related Works}

Automatic ICD coding has been an active research topic in the healthcare domain, with significant research focusing on deep learning approaches. These studies have investigated a range of models, including recurrent neural networks (RNNs), convolutional neural networks (CNNs), graph neural networks (GCNs), and transformer-based models.

CAML\cite{mullenbach-etal-2018-explainable} incorporates multiple CNN-based text encoders and an attention decoder. It is the first attempt to apply a label attention mechanism to the automatic ICD coding task. Several CNN variations have since been developed to tackle the challenges inherent in this problem \cite{luo2021fusion,chen2020towards,ji2020dilated}. EffectiveCAN \cite{liu2021effective}, integrates a squeeze-and-excitation convolution-based network with residual connections, enhancing label attention by leveraging representations from all encoder layers. To address the challenge of long-tail predictions, the authors also incorporated focal loss, trying to improve model performance in rare-label predictions. MultiResCNN \cite{li2020icd} proposes a multi-filter residual convolutional neural network and combines it with a label attention mechanism. LAAT \cite{10.5555/3491440.3491901} utilized a bidirectional Long Short-Term Memory (LSTM) network and integrated it with a customized label-specific attention. PLM-ICD \cite{huang-etal-2022-plm} integrates a pre-trained encoder-based language model with a segment pooling mechanism, which aims to address the challenge of fine-tuning pretrained models with long input texts. These components are then combined with the label attention mechanism originally introduced in LAAT. BERT-XML \cite{zhang2020bert} is a transformer-based model that integrates BERT encoders with multi-label attention mechanisms. Instead of fine-tuning an existing pre-trained BERT model, the authors trained the encoder from scratch using a masked language modeling objective on EHR notes, addressing the challenge of out-of-vocabulary terms. \cite{wang2024multi} proposes a multi-stage retrieval and re-ranking framework. In this approach, for a given clinical note, an initial curated list of ICD codes is predicted, which is then refined through a contrastive learning method to enhance the accuracy of the candidate list. 

Recently \cite{jeong2024bridging} introduced a joint learning framework across medical coding systems, which jointly learns representations from different coding systems through a multi-task learning strategy. \cite{motzfeldt2025code} suggests an agentic framework for medical coding using large language models, which incorporates official coding guidelines designed for human experts and is particularly effective for rare code prediction. Furthermore, \cite{wu2025gom} proposes GoM-ICD, an automatic ICD coding framework that integrates multiple gap schemes within a mixture-of-experts architecture. In addition, \cite{zhang2025general} proposes a general knowledge injection framework that integrates multiple types of ICD-related knowledge into the model.

These research efforts have considerably enhanced the performance of ML-based models for ICD coding. However, none of the prior studies have specifically addressed the order of ICD codes, despite its critical importance in clinical settings.

\begin{figure}[t]
  \includegraphics[width=1.0\linewidth]{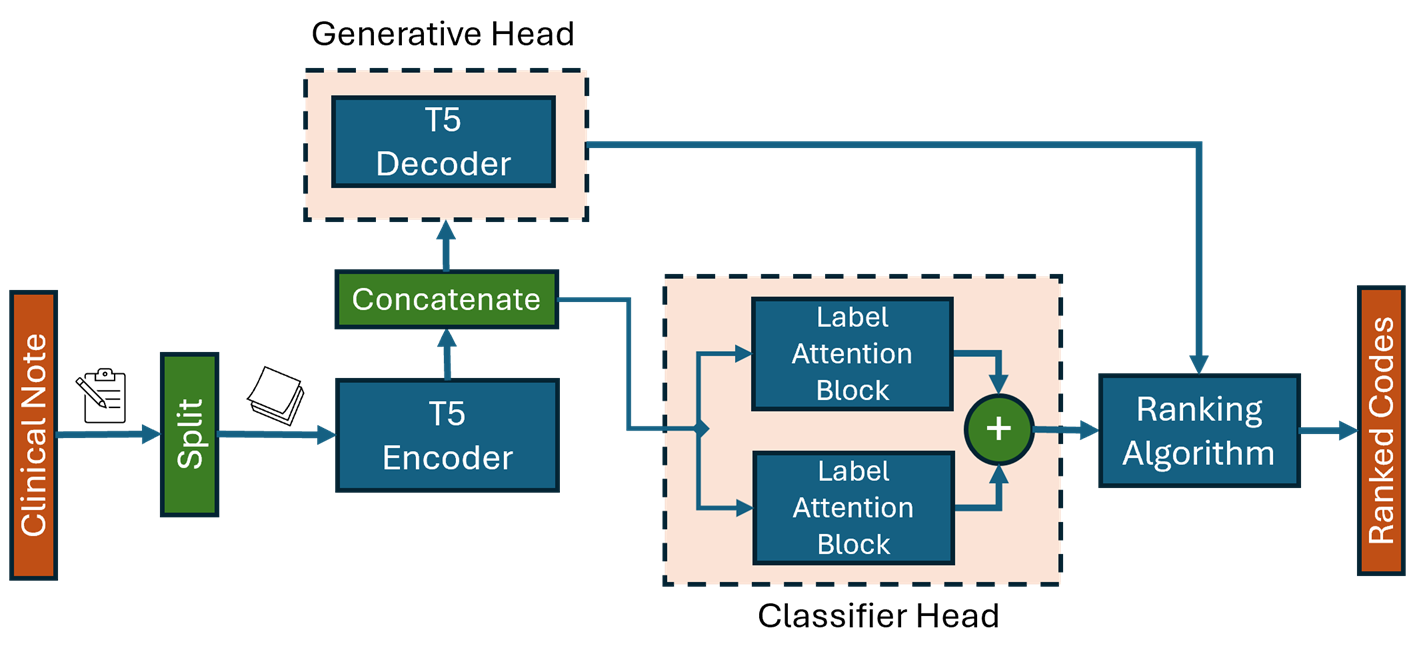}
  \caption{Overview of the proposed LTR-ICD framework for medical code prediction and ranking. The model consists of three main components: a classifier module, a generative module, and a ranking algorithm.}
  \Description{}
  \label{fig:ltricd}
\end{figure}

\section{Methodology}

\subsection{Problem Definition}

ICD coding has traditionally been formulated as an extreme multi-label text classification task \cite{liu2017deep}. Given the fact that the correct sequencing of assigned ICD codes to a clinical note is essential, as a first attempt in this work, we formulate this task as a combined classification and ranking problem. More specifically, given a clinical note, the goal is to generate and recommend an ordered list of ICD codes for the note. To achieve this, we employ a transformer-based pre-trained generative language model and enhance it with a classifier module. The model is trained to jointly learn a set of ICD codes and their corresponding priorities for a given clinical note. Figure~\ref{fig:ltricd} depicts the proposed LTR-ICD (label-to-rank for ICD coding) framework, with details of its components explained in subsequent sections.

\subsection{Pre-trained Language Model}

Using pre-trained language models (PLMs) has become a cornerstone in natural language processing research, consistently driving progress across a multitude of tasks. These models, trained on vast corpora, provide rich contextualized representations of text, allowing for effective adaptation to domain-specific challenges. One of the main challenges of using PLMs for ICD coding is that clinical notes are typically long documents, often exceeding the maximum input length supported by many of these models. To address this, we adopt T5 \cite{raffel2020exploring} as our underlying model. T5 is an encoder-decoder transformer-based language model, which has gained popularity for its unified approach in converting diverse text-based language problems into a standardized text-to-text format. As a generative model, it uses relative positional encoding and can handle input documents of any length, theoretically.

In this work, we modify the standard T5 architecture by incorporating a label attention mechanism and a classification head on top of its encoder block. As a result, our model consists of two complementary components: a classification module that predicts ICD codes without considering their order, and a generative module that produces codes in a sequence reflecting their clinical priority and relevance. Finally, a ranking algorithm integrates the outputs of both components to produce a final ranked list of recommended ICD codes. Detailed descriptions of the classification and generative modules are provided in the Supplementary Material.

\subsection{Ranking Algorithm}

\begin{algorithm}[t]
\renewcommand{\thealgorithm}{\arabic{algorithm}}
\floatname{algorithm}{Algorithm}

\caption{Ranking algorithm combines classifier and generative outputs to produce final ICD predictions}
\label{alg:ranking}

\begin{algorithmic}[1]
    \STATE \textbf{Input:} Predictions of classifier and generative modules
    \STATE \textbf{Output:} LTR-ICD predictions
    \STATE $\text{final\_predictions} \leftarrow \emptyset$
    
    \FOR{each code $c$ in generative\_predictions}
        \IF{$c \in$ classifier\_predictions}
            \STATE append $c$ to final\_predictions
        \ENDIF
    \ENDFOR
    
    \FOR{each code $c$ in classifier\_predictions}
        \IF{$c \notin$ final\_predictions}
            \STATE append $c$ to final\_predictions
        \ENDIF
    \ENDFOR
    
    \STATE \RETURN final\_predictions
\end{algorithmic}

\end{algorithm}

To improve both the ranking characteristics and the classification features of the predicted sequence of ICD codes for a clinical note, we propose a ranking algorithm as depicted in Algorithm~\ref{alg:ranking}. This algorithm integrates the outputs of the classification and generative modules to produce a unified prediction. Through integrating these outputs, the ranking algorithm improves the overall quality of predicted codes and ensures that the sequence of codes aligns more closely with clinical priorities. This combined approach is particularly beneficial in medical scenarios where both the order and accuracy of ICD codes play a critical role in patient care and administrative processes.

\section{Experimental Settings}
\subsection{Dataset}
We trained and evaluated our model on the MIMIC-III dataset \cite{johnson2016mimic}. This dataset is a publicly accessible benchmark database and comprises clinical documents annotated with ICD-9 codes collected from 2001 to 2012. It includes data from about 46,000 patients and contains 15 types of clinical notes. Following the previous studies \cite{yuan2022code,10.5555/3491440.3491901,mullenbach-etal-2018-explainable,10.1609/aaai.v37i4.25668,huang-etal-2022-plm}, we utilized the hospital discharge summaries of the MIMIC-III for ICD coding. In this dataset, each discharge summary is associated with a sequence of ICD codes, including diagnosis and procedure codes. Table~\ref{tbl:stat} shows descriptive statistics for the number of ICD codes assigned to each discharge summary.

Most previous works have used the pipeline provided by \cite{mullenbach-etal-2018-explainable} to pre-process the discharge summaries in MIMIC-III and create the train, test, and validation splits. But \cite{edin2023automated} recently showed that the non-stratified random sampling method used in generating previous splits had been inefficient and resulted in 54\% of the codes in the main dataset not being included in the test split. Subsequently, they introduced new stratified sample splits and reproduced some of the most important previous works using these new datasets. In our work, we use the same splits as in the work of \cite{edin2023automated} and compare our results with their reproduced results for the state-of-the-art classifier model. Additionally, we applied a minimal pre-processing step for model training on the discharge summaries and substituted each de-identification surrogate in these notes with its corresponding entity tag (e.g., [**2151-8-14**] → [DAY], [**Hospital 1708**] → [LOC]).

\begin{table}[t]
\centering
\caption{Frequency statistics of diagnosis and procedure codes per clinical note in the MIMIC-III dataset.}
\begin{tabular}{lll}
\toprule
\textbf{Number of codes}    & \textbf{Diagnosis} & \textbf{Procedure} \\
\midrule
Minimum            & 1         & 0         \\
Maximum            & 39        & 40        \\
Mean               & 11        & 4         \\
Standard deviation & 6.46      & 3.88      \\
Median             & 9         & 3         \\
85th percentile    & 18        & 8         \\
95th percentile    & 24        & 12        \\
\bottomrule
\end{tabular}

\label{tbl:stat}
\end{table}

\subsection{Metrics}
To evaluate the performance of our proposed model, we employ four metrics commonly used in classification tasks, Micro-F1 at K (F1@K), Precision at K (P@K), Recall at K (R@K) and Mean Average Precision at K (MAP@K), as well as one additional metric, Normalized Discounted Cumulative Gain at K (NDCG@K), which is widely used to assess the performance of ranking systems, particularly in information retrieval tasks like search engines and recommender systems \cite{Jaervelin2002}.

It is important to note that, in a typical classification problem, an item in the top-k predictions is considered relevant if it appears in the actual label set. However, this definition of relevance does not offer a way to indicate varying degrees of relevance for the items \cite{borlund2003concept}. As a result, this approach may negatively affect our ability to identify models that can effectively recommend highly relevant items \cite{kekalainen2002using}. To address this and to enable comparison with previous research, we consider an item in the top-k predictions as relevant if it is within the top-k actual labels. This definition of relevance is consistently applied across all the metrics used in this study.

\textbf{Implementation Details:} Detailed implementation settings, training procedures and hyperparameter configurations are provided in the Supplementary Material.

\begin{table*}[t]
    \centering
    \caption{Comparing the three models in terms of different metrics at different ranking positions, K, for diagnosis codes. The best scores between models are indicated in bold.}
    \begin{tabular}{cccccccccccccccc}
        \toprule
        \multicolumn{16}{c}{\textbf{Diagnosis Codes}} \\
        \midrule
        \multicolumn{6}{c}{\textbf{LTR-ICD Model (Ours)}} & 
        \multicolumn{5}{c}{\textbf{GKI-ICD Model}} & 
        \multicolumn{5}{c}{\textbf{PLM-ICD Model}} \\
        \midrule
        @K & F1 & Prec & Rec & MAP & NDCG & 
        F1 & Prec & Rec & MAP & NDCG &
        F1 & Prec & Rec & MAP & NDCG \\
        \midrule
        1  & \textbf{0.474} & \textbf{0.474} & \textbf{0.474} & \textbf{0.474} & \textbf{0.474}  
           & 0.195 & 0.195 & 0.195 & 0.195 & 0.195  
           & 0.202 & 0.202 & 0.202 & 0.202 & 0.202 \\
        2  & \textbf{0.427} & \textbf{0.427} & \textbf{0.426} & \textbf{0.636} & \textbf{0.461}  
           & 0.266 & 0.266 & 0.265 & 0.401 & 0.278  
           & 0.270 & 0.270 & 0.269 & 0.411 & 0.285 \\
        3  & \textbf{0.436} & \textbf{0.437} & \textbf{0.436} & \textbf{0.720} & \textbf{0.485}  
           & 0.333 & 0.335 & 0.331 & 0.543 & 0.353  
           & 0.331 & 0.333 & 0.330 & 0.556 & 0.357 \\
        4  & \textbf{0.448} & \textbf{0.449} & \textbf{0.447} & \textbf{0.753} & \textbf{0.503}  
           & 0.386 & 0.390 & 0.383 & 0.628 & 0.414  
           & 0.379 & 0.382 & 0.377 & 0.642 & 0.415 \\
        5  & \textbf{0.467} & \textbf{0.468} & \textbf{0.465} & \textbf{0.769} & \textbf{0.525}  
           & 0.427 & 0.434 & 0.421 & 0.683 & 0.464  
           & 0.418 & 0.423 & 0.414 & 0.696 & 0.463 \\
        6  & \textbf{0.482} & \textbf{0.484} & \textbf{0.480} & \textbf{0.775} & \textbf{0.542}  
           & 0.462 & 0.472 & 0.452 & 0.721 & 0.504  
           & 0.450 & 0.456 & 0.443 & 0.727 & 0.500 \\
        7  & \textbf{0.496} & 0.500 & \textbf{0.493} & \textbf{0.777} & \textbf{0.557}  
           & 0.487 & \textbf{0.501} & 0.473 & 0.748 & 0.533  
           & 0.476 & 0.486 & 0.467 & 0.751 & 0.530 \\
        8  & \textbf{0.511} & 0.516 & \textbf{0.506} & \textbf{0.776} & \textbf{0.570}  
           & 0.505 & \textbf{0.525} & 0.487 & 0.769 & 0.555  
           & 0.496 & 0.509 & 0.483 & 0.770 & 0.553 \\
        9  & \textbf{0.524} & 0.530 & \textbf{0.518} & 0.776 & \textbf{0.582}  
           & 0.521 & \textbf{0.546} & 0.498 & \textbf{0.785} & 0.572  
           & 0.512 & 0.527 & 0.497 & \textbf{0.785} & 0.572 \\
        10 & \textbf{0.534} & 0.537 & \textbf{0.530} & 0.775 & \textbf{0.591}  
           & 0.533 & \textbf{0.560} & 0.509 & \textbf{0.796} & 0.585  
           & 0.524 & 0.539 & 0.509 & 0.795 & 0.586 \\
        11 & \textbf{0.543} & 0.545 & \textbf{0.540} & 0.774 & \textbf{0.600}  
           & 0.542 & \textbf{0.571} & 0.515 & \textbf{0.806} & 0.595  
           & 0.534 & 0.549 & 0.519 & 0.802 & 0.596 \\
        39 & \textbf{0.578} & 0.584 & \textbf{0.573} & 0.782 & \textbf{0.628}  
           & 0.573 & \textbf{0.634} & 0.523 & \textbf{0.845} & 0.619  
           & 0.569 & 0.603 & 0.538 & 0.837 & 0.626 \\
        \bottomrule
    \end{tabular}
    \label{tbl:diag}
\end{table*}

\begin{table*}[t]
\centering
\caption{Comparing the three models in terms of different metrics at different ranking positions, K, for procedure codes. The best scores between models are indicated in bold.}
    \begin{tabular}{cccccccccccccccc}
        \toprule
        \multicolumn{16}{c}{\textbf{Procedure Codes}} \\
        \midrule
        \multicolumn{6}{c}{\textbf{LTR-ICD Model (Ours)}}  & 
        \multicolumn{5}{c}{\textbf{GKI-ICD Model}} &
        \multicolumn{5}{c}{\textbf{PLM-ICD Model}} \\
        \midrule
        @K & F1 & Prec & Rec & MAP & NDCG &
        F1 & Prec & Rec & MAP & NDCG &
        F1 & Prec & Rec & MAP & NDCG \\
        \midrule
        1  & \textbf{0.572} & \textbf{0.572} & \textbf{0.572} & \textbf{0.572} & \textbf{0.572}  
           & 0.384 & 0.384 & 0.384 & 0.384 & 0.384  
           & 0.425 & 0.425 & 0.425 & 0.425 & 0.425 \\
        2  & \textbf{0.600} & \textbf{0.594} & \textbf{0.605} & \textbf{0.745} & \textbf{0.634}  
           & 0.514 & 0.516 & 0.512 & 0.648 & 0.542  
           & 0.525 & 0.526 & 0.524 & 0.676 & 0.559 \\
        3  & \textbf{0.610} & \textbf{0.601} & \textbf{0.619} & \textbf{0.792} & \textbf{0.663}  
           & 0.582 & 0.588 & 0.576 & 0.746 & 0.620  
           & 0.586 & 0.588 & 0.584 & 0.766 & 0.632 \\
        4  & \textbf{0.620} & 0.609 & \textbf{0.632} & \textbf{0.808} & \textbf{0.682}  
           & 0.617 & \textbf{0.627} & 0.607 & 0.792 & 0.662  
           & 0.616 & 0.620 & 0.612 & 0.801 & 0.669 \\
        40 & 0.681 & 0.663 & \textbf{0.701} & 0.820 & \textbf{0.733}  
           & \textbf{0.686} & \textbf{0.724} & 0.651 & \textbf{0.847} & 0.721  
           & 0.675 & 0.698 & 0.654 & 0.844 & 0.722 \\
        \bottomrule
    \end{tabular}
    
    \label{tbl:proc}
\end{table*}

\section{Experimental Results}
The PLM-ICD model \cite{huang-etal-2022-plm} has demonstrated state-of-the-art performance on the ICD coding classification task using the MIMIC-III-full dataset. However, more recent studies \cite{wang2024multi, wu2025gom, zhang2025general} have reported further improvements over PLM-ICD. Due to the lack of publicly available code or sufficient implementation details for \cite{wang2024multi, wu2025gom}, we were unable to reproduce their results for direct comparison. Therefore, we adopt PLM-ICD and GKI-ICD \cite{zhang2025general} as our primary baseline models. A more comprehensive comparison with other widely used approaches is provided in the Supplementary Material, further demonstrating the effectiveness of the proposed framework.

\subsection{Performance on Evaluation Metrics}
In our study, we compare our findings with those reported by \cite{edin2023automated} for the PLM-ICD classifier. To ensure a fair comparison with previous works, we reproduced the results of the PLM-ICD model using the code provided in \cite{edin2023automated}. For the GKI-ICD model, we used the code provided in \cite{zhang2025general} and trained the model using the data splits introduced in \cite{edin2023automated}. The model was trained for 20 epochs, and we selected the checkpoint with the highest micro-F1 score on the validation set as the best-performing model. The predicted labels for the test data from both models were then ranked according to their logits in descending order to generate a sequence of ICD codes ordered by priority. Since the priorities of diagnosis and procedure codes are not directly comparable within the MIMIC-III dataset, we evaluate and compare the performance of the models separately for diagnosis and procedure codes. This approach allows for a more precise assessment of each model's strengths in handling these distinct categories.

We calculated various performance metrics for the LTR-ICD, GKI-ICD, and PLM-ICD models, with the overall results for diagnosis and procedure codes summarized in Table~\ref{tbl:diag} and Table~\ref{tbl:proc}. As noted in Table~\ref{tbl:stat}, the average number of diagnosis and procedure codes per discharge summary is 11 and 4, respectively. Therefore, the rows corresponding to K=11 and K=4 in Table~\ref{tbl:diag} and Table~\ref{tbl:proc} reflect the models’ performance under typical coding scenarios. Additionally, the maximum number of diagnosis and procedure codes per discharge summary is 39 and 40, respectively, meaning that the rows for K=39 and K=40 represent performance over all available labels for each discharge summary.

\textbf{Micro-F1, Precision and Recall:} The results in Table~\ref{tbl:diag} and Table~\ref{tbl:proc} demonstrate that the LTR-ICD model outperforms the baseline models in detecting top-ranked or high-priority codes in terms of F1@K over all important ranking positions for both diagnosis and procedure codes. In other words, LTR-ICD is more capable of placing high-priority codes at higher ranks than the baseline models. For instance, the precision of our model in identifying the correct primary diagnosis code for a discharge summary is \textasciitilde47\%, while this value is \textasciitilde20\% for the PLM-ICD and GKI-ICD models. Similarly, the LTR-ICD model has a precision of \textasciitilde57\% in detecting the main procedure code. However, this value is \textasciitilde43\% and \textasciitilde38\% for the PLM-ICD and GKI-ICD models, respectively. It is worth noting that as K increases, the results become closer to typical classification metrics due to the binary relevance used in our evaluation. This trend is observed because, at higher ranks, the impact of correctly identifying high-priority codes diminishes, and the evaluation starts reflecting the overall classification performance.

\textbf{MAP and NDCG:} Despite Micro-F1, Precision and Recall, which are micro metrics, MAP and NDCG have a macro nature. Each metric computes scores for individual queries and averages these values across all queries, ensuring equal contribution from each query to the final score. As indicated in Table~\ref{tbl:diag} and Table~\ref{tbl:proc}, these metrics also show the superiority of our proposed model to the baseline models at identifying and prioritizing top-ranked codes. It is important to note that, in NDCG calculations, we assume a binary relevance score for predicted codes. Particularly, for the top-k predicted labels, if a label is in the top-k actual labels, its relevance score would be 1, and 0 otherwise.

Additional comparisons with widely used models using standard classification metrics (F1, Precision, Recall, ROC-AUC, and PR-AUC) are provided in the Supplementary Material.

\subsection{Ranking Capabilities of Models: A Cumulative Gain Approach}
In the context of the MIMIC-III dataset, we encounter a limitation due to the lack of graded relevance scores for the assigned ICD codes to clinical notes. As a result, our evaluation is constrained to binary relevance, where a label is either relevant or not, without intermediate relevance levels. For instance, consider a document with the real ordered labels $[c_1, c_2, c_3]$; if the predictions of two models, A and B, are $[c_1, c_3, c_2]$ and $[c_3, c_2, c_1]$, respectively, both would yield the same NDCG@3 scores despite the evident difference in the order of two predicted sequence of labels and the apparent early ranking performance of model A compared to model B.

To ensure a fair and comprehensive comparison of different models’ performance at a specific ranking position K, we calculate a cumulative gain score for each model up to rank K. This score reflects the average gain achieved up to a given rank, providing a richer picture of how well the models rank the relevant labels at various positions. This approach allows for a more granular and accurate assessment of the ranking capabilities of the models \cite{Jaervelin2002}. We define and compute the cumulative gain (CG) at rank K using the following formula:

\begin{equation}
  \label{eq:eq10}
  {CG}_K=\frac{1}{K}\sum_{k=1}^{K}{NDCG@k}
\end{equation}

\begin{figure}[t]
  \includegraphics[width=\columnwidth]{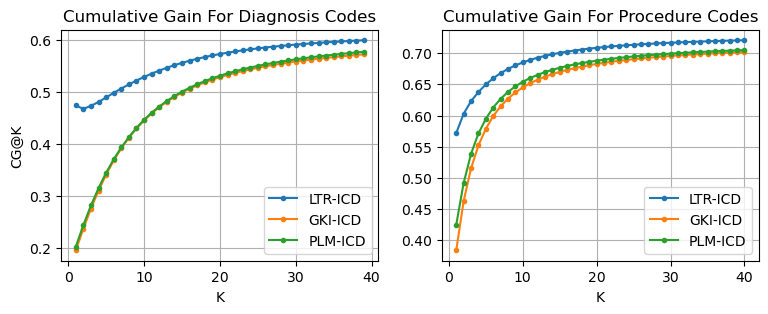}
  \caption{Comparing the performance of the LTR-ICD and baseline models in terms of cumulative gain across multiple ranking positions.}
  \label{fig:cum}
\end{figure}

Figure~\ref{fig:cum} illustrates the throughput of the three models based on this metric. The plots indicate that our proposed model achieves superior early-ranking performance compared to the PLM-ICD and GKI-ICD models and consistently outperforms them for both diagnosis and procedure codes.

\subsection{Ablation Study}
\begin{figure}[t]
  \includegraphics[width=\columnwidth]{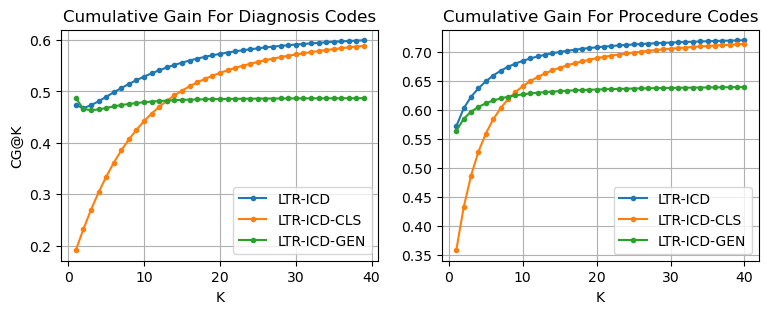}
  \caption{Ablation study comparing classifier-only, generative-only, and full model in terms of cumulative gain across different ranking positions for diagnosis and procedure codes.}
  \label{fig:ablation}
\end{figure}

To better understand the contribution of each component in our framework, we conduct an ablation study comparing three configurations: classifier-only, generative-only, and the full model. We use Cumulative Gain at rank K (CG@K) to evaluate and compare these components. Figure~\ref{fig:ablation} illustrates the performance of these parts for diagnosis and procedure codes, respectively. The results reveal consistent and insightful trends:

\begin{itemize}
    \item The classifier-only module performs reasonably well in identifying relevant codes. However, it shows weaker performance at early ranking positions, highlighting its limitation in modeling code priority and ordering.
    \item The generative-only module achieves stronger performance at early ranks due to its sequential nature, which captures code ordering. However, it suffers from lower coverage, as it tends to miss some relevant codes.
    \item The full model consistently outperforms both individual modules across all ranking positions. This indicates that combining the two components leads to improvements in both ranking quality (especially at early positions) and coverage (overall performance).
\end{itemize}

Overall, these findings confirm that the performance gains of the proposed framework stem from the complementary strengths of the classifier and generative modules, which are effectively integrated through the proposed ranking algorithm.

Further analysis of the effect of label ordering on model performance is presented in the Supplementary Material.

\section{Conclusion}
In this research effort, for the first time, we frame the problem of automatic ICD coding as a classification and ranking assignment and look at this task from a retrieval system point of view. Our proposed LTR-ICD framework demonstrates superior ranking capabilities compared to the state-of-the-art classifier models, particularly in identifying high-priority diagnosis and procedure codes. This work introduces new possibilities for simultaneously learning to predict and rank ICD codes for medical notes.

\begin{acks}
This study was supported by the CAISR Health (Center for Applied Intelligent Systems Research in Health) research profile, funded by the Knowledge Foundation (Grant No. 20200208 01H). 
\end{acks}

\bibliographystyle{ACM-Reference-Format}
\bibliography{bibtxt}

@inproceedings{liu2017deep,
  title={Deep learning for extreme multi-label text classification},
  author={Liu, Jingzhou and Chang, Wei-Cheng and Wu, Yuexin and Yang, Yiming},
  booktitle={Proceedings of the 40th international ACM SIGIR conference on research and development in information retrieval},
  pages={115--124},
  year={2017}
}

@article{raffel2020exploring,
  title={Exploring the limits of transfer learning with a unified text-to-text transformer},
  author={Raffel, Colin and Shazeer, Noam and Roberts, Adam and Lee, Katherine and Narang, Sharan and Matena, Michael and Zhou, Yanqi and Li, Wei and Liu, Peter J},
  journal={Journal of machine learning research},
  volume={21},
  number={140},
  pages={1--67},
  year={2020}
}

@article{vinyals2015order,
  title={Order matters: Sequence to sequence for sets},
  author={Vinyals, Oriol and Bengio, Samy and Kudlur, Manjunath},
  journal={arXiv preprint arXiv:1511.06391},
  year={2015}
}

@article{johnson2016mimic,
  title={MIMIC-III, a freely accessible critical care database},
  author={Johnson, Alistair EW and Pollard, Tom J and Shen, Lu and Lehman, Li-wei H and Feng, Mengling and Ghassemi, Mohammad and Moody, Benjamin and Szolovits, Peter and Anthony Celi, Leo and Mark, Roger G},
  journal={Scientific data},
  volume={3},
  number={1},
  pages={1--9},
  year={2016},
  publisher={Nature Publishing Group}
}

@article{yuan2022code,
  title={Code synonyms do matter: Multiple synonyms matching network for automatic ICD coding},
  author={Yuan, Zheng and Tan, Chuanqi and Huang, Songfang},
  journal={arXiv preprint arXiv:2203.01515},
  year={2022}
}

@article{OMalley2005,
  title = {Measuring Diagnoses: ICD Code Accuracy},
  volume = {40},
  ISSN = {1475-6773},
  url = {http://dx.doi.org/10.1111/j.1475-6773.2005.00444.x},
  DOI = {10.1111/j.1475-6773.2005.00444.x},
  number = {5p2},
  journal = {Health Services Research},
  publisher = {Wiley},
  author = {O’Malley,  Kimberly J. and Cook,  Karon F. and Price,  Matt D. and Wildes,  Kimberly Raiford and Hurdle,  John F. and Ashton,  Carol M.},
  year = {2005},
  month = aug,
  pages = {1620–1639}
}

@article{sutskever2014sequence,
  title={Sequence to sequence learning with neural networks},
  author={Sutskever, Ilya and Vinyals, Oriol and Le, Quoc V},
  journal={Advances in neural information processing systems},
  volume={27},
  year={2014}
}

@article{Jaervelin2002,
  author    = {J{\"a}rvelin, Kalervo and Kek{\"a}l{\"a}inen, Jaana},
  journal   = {ACM Transactions on Information Systems (TOIS)},
  title     = {Cumulated gain-based evaluation of IR techniques},
  year      = {2002},
  number    = {4},
  pages     = {422--446},
  volume    = {20},
  publisher = {ACM New York, NY, USA},
}

@misc{https://doi.org/10.13026/c2xw26,
  url = {https://physionet.org/content/mimiciii/1.4/},
  author = {Johnson,  Alistair and Pollard,  Tom and Mark,  Roger},
  title = {MIMIC-III Clinical Database},
  publisher = {PhysioNet},
  year = {2023}
}

@article{gianfrancesco2021narrative,
  title={A narrative review on the validity of electronic health record-based research in epidemiology},
  author={Gianfrancesco, Milena A and Goldstein, Neal D},
  journal={BMC medical research methodology},
  volume={21},
  number={1},
  pages={234},
  year={2021},
  publisher={Springer}
}

@article{burns2012systematic,
  title={Systematic review of discharge coding accuracy},
  author={Burns, Elaine M and Rigby, E and Mamidanna, R and Bottle, A and Aylin, P and Ziprin, P and Faiz, OD},
  journal={Journal of public health},
  volume={34},
  number={1},
  pages={138--148},
  year={2012},
  publisher={Oxford University Press}
}

@article{borlund2003concept,
  title={The concept of relevance in IR},
  author={Borlund, Pia},
  journal={Journal of the American Society for information Science and Technology},
  volume={54},
  number={10},
  pages={913--925},
  year={2003},
  publisher={Wiley Online Library}
}

@article{kekalainen2002using,
  title={Using graded relevance assessments in IR evaluation},
  author={Kek{\"a}l{\"a}inen, Jaana and J{\"a}rvelin, Kalervo},
  journal={Journal of the American Society for Information Science and Technology},
  volume={53},
  number={13},
  pages={1120--1129},
  year={2002},
  publisher={Wiley Online Library}
}

@inproceedings{10.5555/3491440.3491901,
author = {Vu, Thanh and Nguyen, Dat Quoc and Nguyen, Anthony},
title = {A label attention model for ICD coding from clinical text},
year = {2021},
isbn = {9780999241165},
booktitle = {Proceedings of the Twenty-Ninth International Joint Conference on Artificial Intelligence},
articleno = {461},
numpages = {7},
location = {Yokohama, Yokohama, Japan},
series = {IJCAI'20}
}

@inproceedings{huang-etal-2022-plm,
    title = "{PLM}-{ICD}: Automatic {ICD} Coding with Pretrained Language Models",
    author = "Huang, Chao-Wei  and
      Tsai, Shang-Chi  and
      Chen, Yun-Nung",
    editor = "Naumann, Tristan  and
      Bethard, Steven  and
      Roberts, Kirk  and
      Rumshisky, Anna",
    booktitle = "Proceedings of the 4th Clinical Natural Language Processing Workshop",
    month = jul,
    year = "2022",
    address = "Seattle, WA",
    publisher = "Association for Computational Linguistics",
    url = "https://aclanthology.org/2022.clinicalnlp-1.2",
    doi = "10.18653/v1/2022.clinicalnlp-1.2",
    pages = "10--20"
}

@inproceedings{mullenbach-etal-2018-explainable,
    title = "Explainable Prediction of Medical Codes from Clinical Text",
    author = "Mullenbach, James  and
      Wiegreffe, Sarah  and
      Duke, Jon  and
      Sun, Jimeng  and
      Eisenstein, Jacob",
    editor = "Walker, Marilyn  and
      Ji, Heng  and
      Stent, Amanda",
    booktitle = "Proceedings of the 2018 Conference of the North {A}merican Chapter of the Association for Computational Linguistics: Human Language Technologies, Volume 1 (Long Papers)",
    month = jun,
    year = "2018",
    address = "New Orleans, Louisiana",
    publisher = "Association for Computational Linguistics",
    url = "https://aclanthology.org/N18-1100",
    doi = "10.18653/v1/N18-1100",
    pages = "1101--1111"
}

@InProceedings{Shazeer2018,
  author       = {Shazeer, Noam and Stern, Mitchell},
  booktitle    = {Proceedings of the 35th International Conference on Machine Learning},
  title        = {Adafactor: Adaptive learning rates with sublinear memory cost},
  year         = {2018},
  organization = {PMLR},
  pages        = {4596--4604},
}

@ARTICLE{9705116,
  author={Teng, Fei and Liu, Yiming and Li, Tianrui and Zhang, Yi and Li, Shuangqing and Zhao, Yue},
  journal={IEEE Transactions on Knowledge and Data Engineering}, 
  title={A Review on Deep Neural Networks for ICD Coding}, 
  year={2023},
  volume={35},
  number={5},
  pages={4357-4375},
  doi={10.1109/TKDE.2022.3148267}}

@inproceedings{guo-etal-2022-longt5,
    title = "{L}ong{T}5: {E}fficient Text-To-Text Transformer for Long Sequences",
    author = "Guo, Mandy  and
      Ainslie, Joshua  and
      Uthus, David  and
      Ontanon, Santiago  and
      Ni, Jianmo  and
      Sung, Yun-Hsuan  and
      Yang, Yinfei",
    editor = "Carpuat, Marine  and
      de Marneffe, Marie-Catherine  and
      Meza Ruiz, Ivan Vladimir",
    booktitle = "Findings of the Association for Computational Linguistics: NAACL 2022",
    month = jul,
    year = "2022",
    address = "Seattle, United States",
    publisher = "Association for Computational Linguistics",
    url = "https://aclanthology.org/2022.findings-naacl.55",
    doi = "10.18653/v1/2022.findings-naacl.55",
    pages = "724--736",
}

@article{10.1145/3664615,
author = {Ji, Shaoxiong and Li, Xiaobo and Sun, Wei and Dong, Hang and Taalas, Ara and Zhang, Yijia and Wu, Honghan and Pitk\"{a}nen, Esa and Marttinen, Pekka},
title = {A Unified Review of Deep Learning for Automated Medical Coding},
year = {2024},
publisher = {Association for Computing Machinery},
address = {New York, NY, USA},
issn = {0360-0300},
url = {https://doi.org/10.1145/3664615},
doi = {10.1145/3664615},
journal = {ACM Computing Surveys},
month = {may}
}

@inproceedings{10.1609/aaai.v37i4.25668,
author = {Yang, Zhichao and Kwon, Sunjae and Yao, Zonghai and Yu, Hong},
title = {Multi-label few-shot ICD coding as autoregressive generation with prompt},
year = {2023},
isbn = {978-1-57735-880-0},
publisher = {AAAI Press},
url = {https://doi.org/10.1609/aaai.v37i4.25668},
doi = {10.1609/aaai.v37i4.25668},
booktitle = {Proceedings of the 37th AAAI Conference on Artificial Intelligence and 35th Conference on Innovative Applications of Artificial Intelligence and 30th Symposium on Educational Advances in Artificial Intelligence},
articleno = {599},
numpages = {9},
series = {AAAI'23/IAAI'23/EAAI'23}
}

@inproceedings{lu2022clinicalt5,
  title={ClinicalT5: A generative language model for clinical text},
  author={Lu, Qiuhao and Dou, Dejing and Nguyen, Thien},
  booktitle={Findings of the Association for Computational Linguistics: EMNLP 2022},
  pages={5436--5443},
  year={2022}
}

@article{wang2024multi,
  title={Multi-stage Retrieve and Re-rank Model for Automatic Medical Coding Recommendation},
  author={Wang, Xindi and Mercer, Robert E and Rudzicz, Frank},
  journal={arXiv preprint arXiv:2405.19093},
  year={2024}
}

@inproceedings{sudre2017generalised,
  title={Generalised dice overlap as a deep learning loss function for highly unbalanced segmentations},
  author={Sudre, Carole H and Li, Wenqi and Vercauteren, Tom and Ourselin, Sebastien and Jorge Cardoso, M},
  booktitle={Deep Learning in Medical Image Analysis and Multimodal Learning for Clinical Decision Support: Third International Workshop, DLMIA 2017, and 7th International Workshop, ML-CDS 2017, Held in Conjunction with MICCAI 2017, Qu{\'e}bec City, QC, Canada, September 14, Proceedings 3},
  pages={240--248},
  year={2017},
  organization={Springer}
}

@article{lin2017focal,
  title={Focal Loss for Dense Object Detection},
  author={Lin, T},
  journal={arXiv preprint arXiv:1708.02002},
  year={2017}
}

@article{ji2020dilated,
  title={Dilated convolutional attention network for medical code assignment from clinical text},
  author={Ji, Shaoxiong and Cambria, Erik and Marttinen, Pekka},
  journal={arXiv preprint arXiv:2009.14578},
  year={2020}
}

@inproceedings{li2020icd,
  title={ICD coding from clinical text using multi-filter residual convolutional neural network},
  author={Li, Fei and Yu, Hong},
  booktitle={proceedings of the AAAI conference on artificial intelligence},
  volume={34},
  number={05},
  pages={8180--8187},
  year={2020}
}

@inproceedings{luo2021fusion,
  title={Fusion: Towards automated ICD coding via feature compression},
  author={Luo, Junyu and Xiao, Cao and Glass, Lucas and Sun, Jimeng and Ma, Fenglong},
  booktitle={Findings of the Association for Computational Linguistics: ACL-IJCNLP 2021},
  pages={2096--2101},
  year={2021}
}

@inproceedings{liu2021effective,
  title={Effective convolutional attention network for multi-label clinical document classification},
  author={Liu, Yang and Cheng, Hua and Klopfer, Russell and Gormley, Matthew R and Schaaf, Thomas},
  booktitle={Proceedings of the 2021 Conference on Empirical Methods in Natural Language Processing},
  pages={5941--5953},
  year={2021}
}

@inproceedings{chen2020towards,
  title={Towards interpretable clinical diagnosis with Bayesian network ensembles stacked on entity-aware CNNs},
  author={Chen, Jun and Dai, Xiaoya and Yuan, Quan and Lu, Chao and Huang, Haifeng},
  booktitle={Proceedings of the 58th Annual Meeting of the Association for Computational Linguistics},
  pages={3143--3153},
  year={2020}
}

@article{zhang2020bert,
  title={BERT-XML: Large scale automated ICD coding using BERT pretraining},
  author={Zhang, Zachariah and Liu, Jingshu and Razavian, Narges},
  journal={arXiv preprint arXiv:2006.03685},
  year={2020}
}

@inproceedings{edin2023automated,
  title={Automated medical coding on MIMIC-III and MIMIC-IV: a critical review and replicability study},
  author={Edin, Joakim and Junge, Alexander and Havtorn, Jakob D and Borgholt, Lasse and Maistro, Maria and Ruotsalo, Tuukka and Maal{\o}e, Lars},
  booktitle={Proceedings of the 46th International ACM SIGIR Conference on Research and Development in Information Retrieval},
  pages={2572--2582},
  year={2023}
}

@inproceedings{jeong2024bridging,
  title={Bridging the code gap: A joint learning framework across medical coding systems},
  author={Jeong, Geunyeong and Jeong, Seokwon and Sun, Juoh and Kim, Harksoo},
  booktitle={Proceedings of the 2024 Joint International Conference on Computational Linguistics, Language Resources and Evaluation (LREC-COLING 2024)},
  pages={2520--2525},
  year={2024}
}

@incollection{motzfeldt2025code,
  title={Code like humans: A multi-agent solution for medical coding},
  author={Motzfeldt, Andreas and Edin, Joakim and Christensen, Casper L and Hardmeier, Christian and Maal{\o}e, Lars and Rogers, Anna},
  booktitle={Findings of the Association for Computational Linguistics: EMNLP 2025},
  pages={22612--22627},
  year={2025},
  publisher={Association for Computational Linguistics}
}

@article{wu2025gom,
  title={GoM-ICD: Automatic ICD Coding with Gap Schemes and Mixture of Experts},
  author={Wu, Yifan and Qiu, Weiyan and Zeng, Min and Chen, Xi and Li, Min and Zhu, Hongtao},
  journal={Big Data Mining and Analytics},
  volume={8},
  number={6},
  pages={1211--1224},
  year={2025},
  publisher={TUP}
}

@inproceedings{zhang2025general,
  title={A general knowledge injection framework for ICD coding},
  author={Zhang, Xu and Zhang, Kun and Ma, Wenxin and Wang, Rongsheng and Wu, Chenxu and Li, Yingtai and Zhou, S Kevin},
  booktitle={Findings of the Association for Computational Linguistics: ACL 2025},
  pages={7180--7189},
  year={2025}
}

\appendix
\vspace{25pt}
\noindent
\textbf{\Huge Supplementary Materials}

\section{Methodology}

\subsection{Pre-trained Language Model}

In this work, we modify the standard T5 architecture by incorporating a label attention mechanism and a classification head on top of its encoder block. As a result, our model consists of two complementary components: a classification module that predicts ICD codes without considering their order, and a generative module that produces codes in a sequence reflecting their clinical priority and relevance. Finally, a ranking algorithm integrates the outputs of both components to produce a final ranked list of recommended ICD codes.

\textbf{Classification Module:} The classification module is designed to predict a set of relevant ICD codes for a given clinical note, irrespective of their orders. This component has two main building blocks: an encoder block, which extracts contextual representations from the input clinical text, and a classifier head. The classifier head consists of two label attention blocks, which determine the relative importance of each segment of the input in relation to potential labels.

For the encoder block of the classification module, we utilize the T5 encoder, which is shared between the classification and generative modules. Although the T5 encoder can theoretically process input documents of arbitrary length, to reduce the model's training time and memory usage, we employed the segment pooling mechanism \cite{huang-etal-2022-plm} to extract the hidden representation, $H \in \mathbb{R}^{N \times d_e}$, for an input clinical note. Here, N is the length of the input document and $d_e$ is the encoder’s output embedding dimension.

Then, the text representation, $H$, is fed into each label attention block to obtain the label-wise attention matrix $A$. Attention weights are calculated in a two-stage convolutional process using two convolutional filters $W_{1c}\in\mathbb{R}^{k\times d_e\times d_c}$ and $W_{2c}\in\mathbb{R}^{k\times d_c\times L}$, where $k$ is the kernel size, $d_c$ is the first filter’s output size and $L$ is the total number of labels. Our proposed label attention mechanism is the following,

\begin{equation}
  \label{eq:eq1}
  Z=tanh\left(Conv\left(W_{1c},H\right)\right)
\end{equation}
\begin{equation}
  \label{eq:eq2}
  A=Softmax\left(Conv\left(W_{2c},Z\right)+P\right)
\end{equation}

The inputs to the filters are padded, so that $Z\in\mathbb{R}^{N\times d_c}$ and $A\in\mathbb{R}^{N\times L}$. For the $n$-th token of the input text, $z_n$ is calculated as

\begin{equation}
  \label{eq:eq3}
  z_n=tanh\left(W_{1c}\ast H_{n:n+k}\right)
\end{equation}

where $H_{n:n+k}\in\mathbb{R}^{k\times d_e}$ and * is the convolution operator.

In our attention mechanism, we introduce the parameter $P\in\mathbb{R}^{N\times 1}$ as a position-aware bias. Since the segment pooling mechanism splits the input text into separate segments before computing hidden representations, standard positional encodings may not be effectively preserved. To address this, we incorporate P as an additional learnable bias term in the attention computation, ensuring that the model retains positional information when computing attention scores.

Then, we utilize the attention weights, $A$, and calculate the label-based document representation $R\in\mathbb{R}^{L\times d_e}$,

\begin{equation}
  \label{eq:eq4}
  R=A^TH
\end{equation}

Finally, by applying a linear layer to $R$, we obtain labels’ logits related to the attention block.

\begin{equation}
  \label{eq:eq5}
  l^b=LL\left(R\right)
\end{equation}

Our classifier head features two label attention blocks, whose computed label logits are combined through summation.

\textbf{Generative Module:} The generative module is designed to generate an ordered sequence of ICD codes for a given clinical note, ranking the codes from the most to the least relevant. While all generated codes may be relevant, the module prioritizes them based on their importance to the note.

The generative module leverages the hidden representations produced by the shared encoder through its cross-attention mechanism, enabling it to condition the generation process on the full contextual understanding of the input note. Given an input text T and previously generated output tokens $t_1, t_2, \ldots, t_{i-1}$, this module calculates the probability of the next token $t_i$ at the output. This step is repeated multiple times until the model generates the end-of-sequence token at step L. Finally, all generated tokens are grouped as a sequence of ICD codes, $C$, for the input text, $T$.

\begin{equation}
  \label{eq:eq6}
  Pr\left(C|T\right)=\prod_{i=1}^{L}Pr{\left(t_i\middle| t_1,\ldots,t_{i-1},T\right)}
\end{equation}

Since ICD codes are generated sequentially, the model learns to rank them based on their importance for a given clinical note during training. In other words, this sequential approach enables the module to capture the inherent hierarchy and relevance of ICD codes.

\section{Implementation details}
\subsection{Training}
In our study, we modify the standard T5 architecture by adding a custom classification head on top of its encoder. To efficiently process long clinical notes, we also split the input text into fixed-length segments before feeding them into the encoder. The encoder processes each segment individually, and the resulting hidden representations are concatenated at the encoder’s output. This combined representation is then used by both the classification head and the decoder’s cross-attention mechanism. The model is trained to jointly learn two tasks: predicting ICD codes through the classification module and generating them in order of importance via the generative module. The following section describes the training setup for these components.

\textbf{Label Processing for Generative Module:} In this work, predicting both diagnosis and procedure codes is formulated as a single task. For each discharge summary, we order its related ICD codes (diagnosis and procedure codes) by their sequence numbers (SEQ\_NUM), as they exist in the DIAGNOSES\_ICD and PROCEDURES\_ICD tables in the MIMIC-III dataset \cite{https://doi.org/10.13026/c2xw26}. Sequence numbers are ordinal labels assigned by expert coders to associated ICD codes to an admission, showing the priority of the codes for that specific admission. We use semicolons to separate codes within a sequence of labels. Since the ordering of the labels could significantly impact the model’s performance \cite{vinyals2015order}, we experimented with three different settings for arranging diagnosis and procedure codes. Particularly, given the code set \{diagnosis: $[d_1; d_2; d_3]$, procedure: $[p_1; p_2]$\} for a sample admission, the three tested code orderings are as follows,

\begin{enumerate}
    \item Ordered diagnosis codes followed by ordered procedure codes, i.e. $[d_1; d_2; d_3; p_1; p_2]$.
    \item Ordered procedure codes followed by ordered diagnosis codes, i.e. $[p_1; p_2; d_1; d_2; d_3]$.
    \item Mixed ordering of diagnosis and procedure codes based on priority from high to low, i.e. $[d_1; p_1; d_2; p_2; d_3]$
\end{enumerate}

We trained and evaluated three different models using each of the orderings above. As indicated in section 6.4, our experiments demonstrate that the third ordering (combining diagnosis and procedure codes based on priority) produced the best results compared to the first two. Consequently, we adopted this format for the sequence of target labels in our generative module.

\textbf{Model Configurations and Training:} In our experiment, we utilized ClinicalT5 \cite{lu2022clinicalt5}, a T5-based domain-specific language model that is specifically designed for biomedical and clinical text processing. We initialized our model using the pretrained Clinical-T5-Base weights provided by PhysioNet\footnote{\url{https://www.physionet.org}}.
To ensure consistent input formatting, clinical notes were either truncated or padded to a fixed length of 5120 tokens. For the segment pooling mechanism, each segment is set to a length of 512 tokens. For the classification module, output labels were represented using multi-hot encoding, and for the generative module, the maximum output sequence length was set to 256 tokens.

During training, the Adafactor optimizer \cite{Shazeer2018} was employed for optimization, and the batch size was set to 6. The training was conducted in a two-phase procedure. In the first phase, the model was trained jointly using Focal Loss \cite{lin2017focal} for the classification module and cross-entropy loss for the generative module. The total loss for this phase, $L_1$, is defined as:

\begin{equation}
  \label{eq:eq7}
  L_1=\ L_c+\alpha L_g
\end{equation}

Where $L_c$ and $L_g$ denote the losses for the classification and generative heads, respectively. In our experiment, the weighting coefficient $\alpha$ was set to 0.02, and the Focal loss parameter $\gamma$ was set to 2. For this phase, the model with the highest micro-F1 score on the validation data is selected as the best-performing model.

In the second phase, we freeze the encoder and decoder components of the best model obtained from the first phase and continue training only its classifier head using Dice loss \cite{sudre2017generalised}, defined as:

\begin{equation}
  \label{eq:eq8}
  L_2=1-\frac{2\sum_{1}^{N}{y_i\left(\sigma\left(l_i^{\left(b1\right)}\right)+\sigma\left(l_i^{\left(b2\right)}\right)\right)}}{\sum_{1}^{N}\left(y_i+\sigma\left(l_i^{\left(b1\right)}\right)+\sigma\left(l_i^{\left(b2\right)}\right)\right)}
\end{equation}

Here, $l_i^{\left(b1\right)}$ and $l_i^{\left(b2\right)}$ represent the logits of the attention blocks, $y_i\in\left\{0,\ 1\right\}$ denotes to the ground truth label, and N is the total number of labels. Similar to the first phase, we monitor the micro-F1 score on the validation data and choose the model with the highest score as the final model. The learning rates for the first and second phases are set to 0.001 and 0.0001, respectively \cite{guo-etal-2022-longt5}.

\subsection{Inference}
At inference time for the generative module, we utilize the beam search strategy \cite{sutskever2014sequence} to generate a sequence of codes for the input note. Due to memory and time constraints, we limit the beam size to a maximum of 5. Then, the generated sequence is split by a separating token, resulting in a list of labels. Since generative language models could generate repetitive outputs and are prone to hallucination, we further post-process the generated labels and remove repeated labels and those not ICD codes.

\begin{table*}[t]
\centering
\caption{Comparing LTR-ICD model with previous widely used models across different classification metrics. The best scores among models are indicated in bold. \textsuperscript{+} indicates a statistically significant improvement over the baseline (paired bootstrap test, 10,000 resamples; all p < 0.002).}
    \begin{tabular}{lcccccccccc}
    \toprule
    \multirow{2}{*}{} & \multicolumn{2}{c}{F1} & \multicolumn{2}{c}{Precision} & \multicolumn{2}{c}{Recall} & \multicolumn{2}{c}{AUC-ROC} & \multicolumn{2}{c}{AUC-PR} \\
    \midrule
    Model                  & Micro      & Macro     & Micro         & Macro         & Micro        & Macro       & Micro        & Macro        & Micro        & Macro       \\
    \midrule
    Bi-GRU            & 49.51      & 11.29     & 54.73         & 14.99         & 45.20        & 10.78       & 97.62        & 90.15        & 48.14        & 15.82       \\
    CAML              & 55.40      & 20.71     & 55.97         & 22.53         & 54.84        & 21.49       & 98.05        & 89.96        & 55.41        & 23.31       \\
    MultiResCNN       & 55.93      & 23.86     & 55.19         & 24.70         & 56.68        & 25.98       & 98.12        & 91.41        & 56.00        & 26.45       \\
    LAAT              & 57.47      & 21.79     & 62.61         & 26.93         & 53.11        & 21.20       & 98.49        & 93.16        & 58.73        & 28.84       \\
    PLM-ICD           & 59.73      & 26.60     & 62.91          & 30.56         & 56.86        & 26.61       & 98.85         & 95.05         & 61.89         & 33.95       \\
    GKI-ICD           & 60.35      & 27.41     & \textbf{65.97}          & \textbf{32.86}         & 55.61        & 27.24       & \textbf{99.03}         & \textbf{96.68}         & \textbf{62.95}         & \textbf{37.48}       \\
    \midrule
    LTR-ICD   (Ours)  & \textbf{60.65}\textsuperscript{+}      & \textbf{29.04}\textsuperscript{+}     & 60.57         & 32.60         & \textbf{60.74}\textsuperscript{+}        & \textbf{29.54}\textsuperscript{+}       & 98.20        & 93.27        & 60.47        & 34.90       \\
    \bottomrule
    \end{tabular}
    
    \label{tbl:appx}
\end{table*}

\section{Experimental Results}
\subsection{Extended Classification Performance Comparison}
The primary focus of this work is on both classification and ranking for ICD coding, a novel formulation that, to the best of our knowledge, has not been previously explored in literature. Accordingly, in the previous section, we compared our proposed model primarily with the state-of-the-art PLM-ICD and GKI-ICD models in terms of both ranking and classification performance. In this section, we provide an additional evaluation that focuses exclusively on classification performance.
We compare our model with a selected set of widely used ICD coding models using standard classification metrics: F1 score, precision, recall, ROC-AUC, and PR-AUC. We report both micro and macro values for each of these metrics to provide a comprehensive evaluation of classification performance. Models without publicly available source code were excluded from this comparison, as their results could not be reliably reproduced due to missing implementation details. Additionally, models utilizing multi-modal inputs, such as code descriptions, synonyms, or hierarchical structures, were excluded due to their added complexity and lack of clear evidence for significant performance improvements \cite{edin2023automated}. Therefore, for this evaluation, we selected four representative models: Bi-GRU \cite{mullenbach-etal-2018-explainable}, CAML \cite{mullenbach-etal-2018-explainable}, MultiResCNN \cite{li2020icd}, and LAAT \cite{10.5555/3491440.3491901}, along with PLM-ICD and GKI-ICD as strong state-of-the-art baselines. Table~\ref{tbl:appx} presents the classification performance of these models alongside our proposed LTR-ICD framework. The results highlight the effectiveness and robustness of our approach from a purely classification standpoint, independent of ranking considerations. Notably, our model achieves substantial improvements over prior methods in multiple metrics. Specifically, our model could improve the previous state-of-the-art macro-F1 score from 27.41 to 29.04, underscoring its capacity to better handle label imbalance and rare codes.

\subsection{Impact of Label Ordering on Model Performance}

\begin{figure}[t]
  \includegraphics[width=\columnwidth]{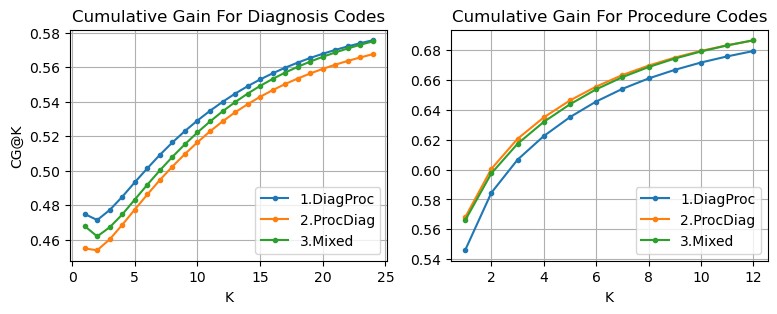}
  \caption{Comparing the impact of diagnosis and procedure codes ordering on the performance of the LTR-ICD model, measured by cumulative gain at K using test data.}
  \label{fig:order}
\end{figure}

As demonstrated by \cite{vinyals2015order}, the order in which target labels are presented to a generative model can significantly impact its learning process and overall performance. In our study, we explored this phenomenon in the context of ICD coding. We trained three distinct models, each utilizing a different ordering of diagnosis and procedure codes for the generative module, as outlined previously. The resulting models were evaluated using the cumulative gain approach, with the outcomes depicted in Figure~\ref{fig:order}. Our experiments revealed that the third ordering, which combines diagnosis and procedure codes based on their priority, yielded the best performance across both code types. This performance gain could be attributed to the generative module’s ability to learn more meaningful representations and make more accurate predictions by prioritizing codes based on their clinical relevance. Furthermore, comparing the blue and orange lines in Figure~\ref{fig:order}, one can clearly observe that a model trained with the first ordering (diagnosis codes followed by procedure codes) performs better in predicting diagnosis codes. Conversely, a model trained with the second ordering (procedure codes followed by diagnosis codes) better predicts procedure codes, which align with our expectations.

\end{document}